\documentclass[preprint,12pt,authoryear]{elsarticle}

\usepackage[T1]{fontenc}
\usepackage[utf8]{inputenc}
\usepackage{amsmath,amssymb}
\usepackage{booktabs}
\usepackage{array}
\usepackage{threeparttable}
\usepackage{graphicx}
\usepackage{url}
\usepackage[hidelinks]{hyperref}

\journal{International Journal of Applied Earth Observation and Geoinformation}

\makeatletter
\def\ps@pprintTitle{%
  \let\@oddhead\@empty
  \let\@evenhead\@empty
  \let\@oddfoot\@empty
  \let\@evenfoot\@empty}
\makeatother

\begin{document}

\begin{frontmatter}

\title{DamageArbiter: A Multimodal Arbitration Framework for Disaster Damage Assessment from Street-View Imagery}

\author[tamugeo]{Yifan Yang}
\author[tamugeo]{Lei Zou\corref{cor1}}
\ead{lzou@tamu.edu}
\author[tamulaup]{Wenjing Gong}
\author[ufise]{Kani Fu}
\author[tamugeo]{Zongrong Li}
\author[usc]{Siqin Wang}
\author[utk]{Bing Zhou}
\author[tamugeo]{Heng Cai}
\author[tamugeo]{Hao Tian}

\cortext[cor1]{Corresponding author.}

\affiliation[tamugeo]{organization={Department of Geography, Texas A\&M University},
            city={College Station},
            country={USA}}
\affiliation[tamulaup]{organization={Department of Landscape Architecture \& Urban Planning, Texas A\&M University},
            city={College Station},
            country={USA}}
\affiliation[ufise]{organization={Department of Industrial and Systems Engineering, University of Florida},
            city={Gainesville},
            country={USA}}
\affiliation[usc]{organization={Spatial Sciences Institute, University of Southern California},
            city={Los Angeles},
            country={USA}}
\affiliation[utk]{organization={Department of Geography and Sustainability, University of Tennessee},
            city={Knoxville},
            country={USA}}

\begin{abstract}
Analyzing street-view imagery with computer vision models offers a promising approach for rapid, hyperlocal disaster damage assessment, but existing approaches typically rely on ``black box'' pre-trained vision models, which lack interpretability and reliability. This study proposes DamageArbiter, a multimodal disagreement-driven arbitration framework designed to improve the accuracy and reliability of street-view-based damage assessment. DamageArbiter leverages the complementary strengths of unimodal and multimodal models and employs a lightweight logistic regression meta-classifier to arbitrate cases in which model predictions disagree. Using 2,556 post-disaster street-view images, paired with manually generated or large language model (LLM)-generated text descriptions, we systematically compared DamageArbiter with fine-tuned unimodal models, including image-only and text-only models, and CLIP-based multimodal models in terms of classification performance and overconfidence errors. Results show that DamageArbiter improved accuracy to 75.85\% and the Matthews correlation coefficient (MCC) to 0.6188, compared with the best-performing text-only baseline (63.07\% accuracy, 0.4126 MCC), image-only baseline (74.33\% accuracy, 0.5947 MCC), and CLIP baseline (74.22\% accuracy, 0.5915 MCC). The overconfidence analysis further reveals that DamageArbiter substantially reduced the overconfidence error from 70.58\% for the best-performing baseline model, the image-only ViT model, to 16.45\%. Overall, this study demonstrates that accuracy alone is insufficient for evaluating disaster damage classification models and highlights the importance of measuring overconfidence errors as part of model reliability assessment. By improving prediction accuracy while significantly reducing overconfident errors, DamageArbiter offers a more reliable framework for rapid, hyperlocal disaster damage assessment from street-view imagery.

\end{abstract}


\end{frontmatter}
\section{Introduction}

Timely and accurate post-disaster damage assessments play a pivotal role in supporting emergency response, guiding resource allocation, and informing short-term and long-term recovery efforts. Among the diverse data sources employed in disaster damage assessment, street-view imagery has emerged as a valuable source for high-resolution, human-perspective observations of structural and infrastructural damages that are often difficult to capture from aerial or satellite imagery \citep{ref12,ref38,ref37}. For example, post-disaster street-view images can reveal fine-scale damage indicators, such as malfunctioned utility poles, building watermarks, inundated streets, debris, damaged roadways, and fallen trees. These visual signals provide critical evidence for assessing localized impacts of disasters. Consequently, there is growing interest in both research and practice in leveraging street-view imagery for post-disaster damage assessment.

Traditional approaches to street-view-based damage assessments typically use manually labeled datasets to train machine learning models or fine-tune pre-trained computer vision models, such as the Visual Geometry Group network (VGG), the Shifted Window Transformer (Swin Transformer), and ConvNeXt, to classify disaster impacts into different levels \citep{ref13,ref36,ref42}. However, several critical limitations exist. First, these methods rely on extensive manual annotations, which are costly and time-consuming, making them slow to adapt to new or evolving disaster scenarios. Second, most models are evaluated primarily using conventional classification metrics, such as accuracy or F1-score, which provide limited insight into model reliability. In disaster contexts, this is a major concern because an incorrect prediction made with high confidence may be more problematic than an uncertain prediction, especially when model outputs are used to support rapid situational awareness, resource allocation, or field prioritization. Third, existing approaches often rely on a single model or a single data modality, making them vulnerable to modality-specific errors. Image-only models may capture visual damage patterns effectively but can produce overconfident errors when visual evidence is ambiguous.

Recent advances in vision-language models (VLMs) open promising avenues for disaster damage assessment using street-view imagery. By aligning images with descriptions, VLMs can capture richer semantics from images, supporting fine-grained reasoning and classifications. In particular, the Contrastive Language-Image Pretraining (CLIP) model provides a general framework for cross-modal alignment that can be adapted for disaster scenarios. \citet{ref41} demonstrated the potential of VLMs in perceiving multidimensional disaster impacts from street-view images. However, the application of VLMs in disaster damage assessment remains in its early stages, primarily due to the lack of high-quality disaster-related image-text pair datasets for fine-tuning and benchmarking VLM models \citep{ref11,ref28}. Further research is needed to examine whether VLMs can enhance street-view-based damage assessment and to systematically evaluate their accuracy, reliability, and decision-support utility.

This research proposes DamageArbiter, a disagreement-driven multimodal arbitration framework for damage assessment from street-view imagery. The overarching objective is to develop the DamageArbiter framework and benchmark its accuracy and reliability against baseline unimodal models, including text-only or image-only approaches, as well as multimodal CLIP models. In this work, model reliability is evaluated through overconfidence, defined as the systematic tendency for a model to assign high confidence to its predictions even when those predictions are incorrect or uncertain. Post-disaster street-view images collected after the 2024 Hurricane Milton, along with textual descriptions generated by human annotators and large language models (LLMs), were used to fine-tune the models and benchmark their performance. LLM-generated descriptions are compared with human annotations to evaluate whether automated descriptions can support robust disaster damage assessment. We validated the models' effectiveness and deployed the optimal configuration to geo-visualize street-level damages in the Horseshoe Beach, Florida, after Hurricane Milton. By explicitly measuring overconfidence and using model disagreement as an arbitration signal, this study demonstrates that accuracy alone is insufficient for evaluating disaster damage classification models and contributes a more accurate and reliability-aware framework for rapid, hyperlocal damage assessment from street-view imagery.

\section{Related Work}

\subsection{Analysis of Street-View Imagery for Disaster Damage Assessment}

Street-view imagery has become an increasingly valuable data source for fine-grained, human-perspective urban analytics. These images can capture street-level details, such as building characteristics, the availability and quality of green space, and infrastructure conditions. Platforms such as Mapillary, Google Street View, and OpenStreetMap (OSM) enable widespread access to such data at different locations and timestamps, facilitating large-scale, hyperlocal, and longitudinal urban analysis at unprecedented scales \citep{ref2,ref22,ref25}. Meanwhile, advances in deep learning-based computer vision have significantly accelerated research in visual-based investigations. Rather than training models from scratch, researchers have fine-tuned pre-trained visual backbones, such as VGG, Swin Transformer, and ConvNeXt, leading to notable gains in image-based classification accuracy and computational efficiency \citep{ref9,ref3,ref34}. Among them, Swin Transformer leverages hierarchical attention mechanisms to capture both local and global visual patterns \citep{ref20,ref19}, while ConvNeXt employs large-kernel convolutions to enhance the representational capacity of images beyond traditional convolutional neural networks (CNNs) \citep{ref4,ref31}.

Recent studies have begun leveraging pre-trained vision models for disaster damage assessment tasks \citep{ref22,ref36,ref43}. For example, \citet{ref5} employed a YOLOv5-based computer vision framework to estimate First-Floor Elevation (FFE) from street-view images, providing a scalable alternative to field surveys and supporting flood mitigation planning in coastal regions. \citet{ref27} adopted a Mask Region-based CNN (Mask R-CNN) approach to extract visual features related to building structure, condition, and surrounding infrastructure, and used these indicators as proxies to estimate household-level vulnerability to natural hazards. \citet{ref36} fused street-view imagery with structured building information to assess post-hurricane building damage. \citet{ref13} proposed CVDisaster, a cross-view framework that combines street-view and very-high-resolution satellite imagery for mapping hurricane damage. Their framework achieved approximately 80\% accuracy in geolocating disaster-related street-view images and 75\% accuracy in assessing damage from those images. \citet{ref40} further showed that bi-temporal street-view imagery can improve hyperlocal damage estimation, with a dual-channel Feature-Fusion ConvNeXt model increasing accuracy from 66.14\% for the Swin Transformer baseline to 77.11\%.

\subsection{Analyzing Street-View Imagery with Vision-Language Models (VLMs)}

VLMs are multimodal models that jointly learn from visual and textual data to support cross-modal understanding, reasoning, and downstream applications. Conventional VLMs usually align image and text encoders with paired image-text data, using contrastive objectives to pull matched pairs closer in a shared embedding space and push mismatched pairs apart \citep{ref24}. Newer generative VLMs often connect a vision encoder to an LLM and refine the combined model through instruction tuning or supervised multimodal conversations \citep{ref14,ref17,ref23}. Commercial systems include GPT-4o and later versions, as well as Gemini, while open-source or research-oriented examples include CLIP, BLIP-2, and LLaVA \citep{ref6,ref14,ref17,ref23,ref24}. Because of VLMs' capacity in connecting visual evidence with language representations, they can support tasks such as image-text retrieval, captioning, visual question answering, classification, and semantic reasoning.

Among VLMs, CLIP has become especially influential. CLIP learns aligned visual and textual representations through contrastive pre-training \citep{ref24}. CLIP-inspired methods have been applied to open-domain image geolocation \citep{ref10}, urban image retrieval and reasoning \citep{ref15}, and cross-view disaster mapping with street-view and satellite imagery \citep{ref13}. Previous work also enriched images with LLM-generated texts and input them into CLIP for downstream tasks, including street-view-based disaster impact investigations. For example, \citet{ref11} paired ground-level building images with their corresponding ChatGPT-generated captions using VLMs, including CLIP and ViLT (Vision-Language Transformers), to assess post-disaster building damages. They reported that the CLIP model reached a validation macro-F1 score of 0.61, outperforming image-only ViT-B/32 (0.57) and text-only models (0.52). \citet{ref16} proposed CLIP-BCA-Gated (CLIP with Bi-Cross-Attention and Adaptive Gating) to classify multimodal disaster-related Twitter/X messages, achieving 91.77\% accuracy and 91.74\% F1 on CrisisMMD (\url{https://crisisnlp.qcri.org/crisismmd.html}), a Multimodal Crisis Dataset.

These CLIP-related studies demonstrate the value of image-text alignment for disaster and crisis analysis, but they also leave important gaps. Most report classification performance without systematically examining the causes of erroneous predictions. In addition, many studies use automatically generated captions as semantic input without validating them against human-written descriptions or testing whether generated descriptions help or confuse model decisions. Furthermore, the reliability of CLIP-based damage assessment remains underexplored. Modern neural networks can be mis-calibrated, assigning high confidence to incorrect predictions even when uncertainty is warranted \citep{ref8}. Overconfidence errors are incorrect predictions made with a large confidence margin, in which the model assigns a high probability to the wrong class. Ambiguity errors are low-margin cases in which the probabilities for competing severity classes are similar or where visual and textual evidence conflict. These concepts capture distinct reliability risks: overconfidence reflects false certainty, while ambiguity reflects unstable evidence. Both gaps are central to image-based disaster damage assessment, where visual cues can be subtle, partially occluded, or semantically ambiguous. The DamageArbiter model developed by this study is poised to address the aforementioned limitations and challenges.

\section{Data and Study Area}

The Milton street-view imagery dataset (Milton-SV) was used for the experiments. It was collected after Hurricane Milton in 2024 in Horseshoe Beach, Florida, United States, and was constructed by \citet{ref40}. The dataset contains 2,556 post-disaster street-view images that were manually classified into three damage severity levels: mild (658 images), moderate (1,196 images), and severe (702 images). Figure 1 shows the locations and damage labels of each street-view image included in the Milton-SV dataset. The manual labeling process was conducted by multiple trained annotators using predefined criteria based on observable disaster impacts, including fallen trees, building debris, damaged infrastructure, and inundated roads. This human annotation closely mirrors the way humans perceive and interpret disaster scenes and aligns with CLIP's core objective of learning image-text semantic alignment. A cross-validation process was subsequently performed to ensure annotation consistency and accuracy.

\begin{figure}[htbp]
\centering
\includegraphics[width=0.95\textwidth]{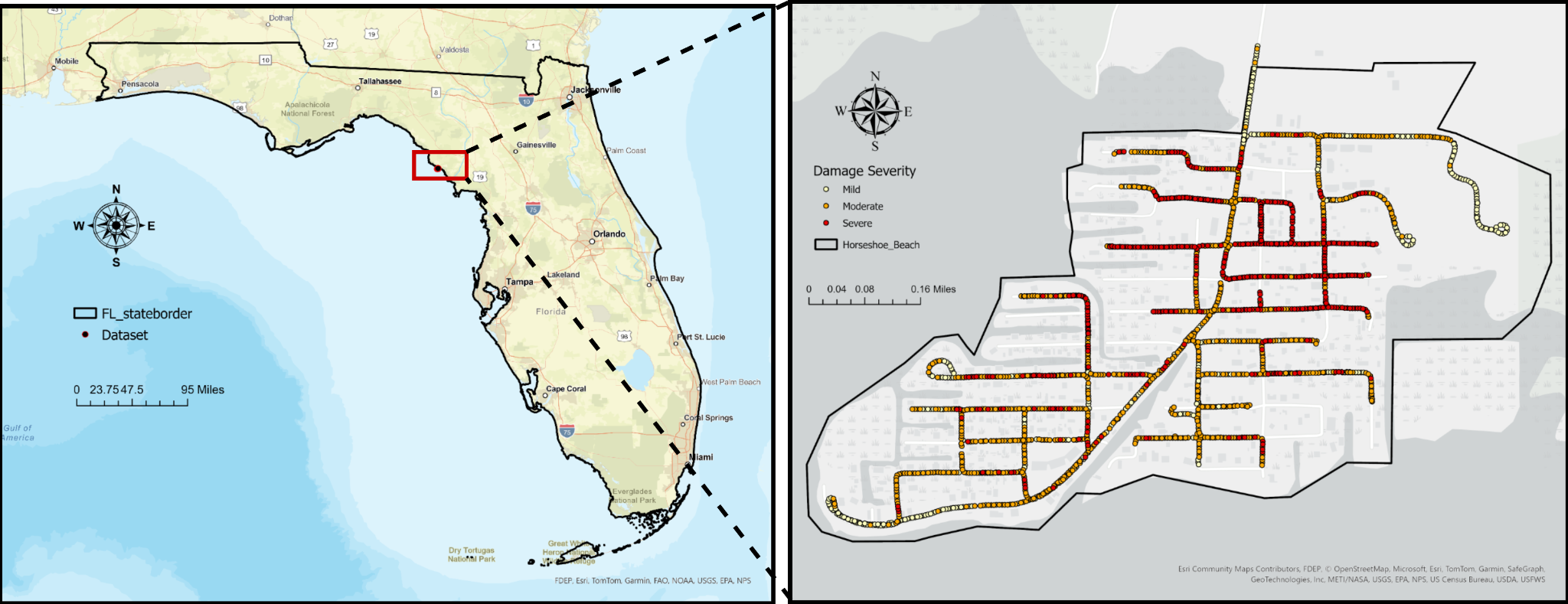}
\caption{Hurricane Milton Damage Mapping and Street-View Sample Distribution}
\label{fig:damage-mapping}
\end{figure}

In parallel, we conducted a fine-grained descriptive annotation to enhance the interpretability and semantic depth of disaster assessment using automated annotation and human captioning. For automated annotation, we tested two cost-effective LLMs, GPT-4o-mini and Gemini 2.0 Flash-Exp, given their ability to balance annotation quality, inference speed, and computational cost for large-scale image labeling tasks. The automated annotation process was guided by a structured prompt designed to constrain the model's behavior and reduce hallucinations. Specifically, the model was instructed to act as a disaster-damage annotator and describe only damage evidence explicitly visible in each image. The prompt emphasized factual, image-grounded descriptions and prohibited speculative or inferred content. The model was required to generate a concise damage description of approximately 50 words, focusing on observable damages, including fallen trees, damaged buildings, debris accumulation, and flooding conditions. All outputs were constrained to a predefined JSON schema to ensure structural consistency and facilitate downstream classification. The generated captions from each LLM were compared with the original street-view images using CLIPScore \citep{ref24}, and the higher-scoring outputs were selected as the textual dataset for the subsequent experiments. GPT-4o-mini produced the higher-scoring outputs and was therefore used in the LLM-caption setting for the experiments.

After descriptive annotation, each street-view image sample is paired with human- and LLM-generated text descriptions, enabling a systematic comparison of manual and automated perception of the same visual scene (Figure 2). This dataset design supports robust classification experiments, semantic interpretation, and evaluation of textual data's role in disaster assessment.

\begin{figure}[htbp]
\centering
\includegraphics[width=0.95\textwidth]{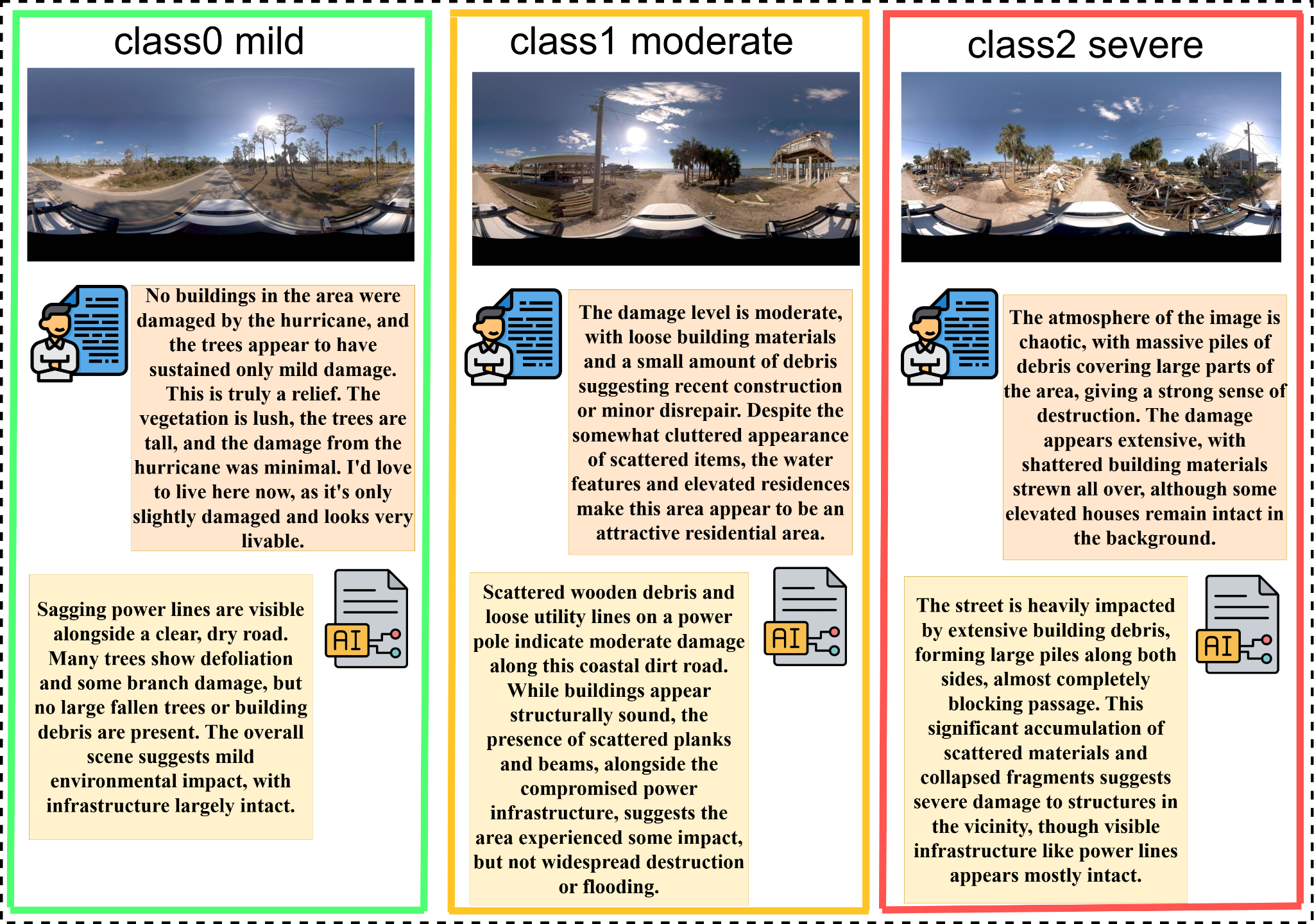}
\caption{Examples of human- and LLM-generated descriptions of street-view images across different damage severity levels}
\label{fig:caption-examples}
\end{figure}

\section{Methodology}

\subsection{Experimental Design}

The study compared the performance of four architectures for classifying disaster damages using street-view images and fine-tuned pre-trained models (Figure 3). These architectures gradually expand street-view-based disaster damage assessment tasks from unimodal feature extraction to multimodal alignment and disagreement-driven arbitration, aiming to improve overall reliability and performance. Three baseline architectures were designed based on input data modality: image unimodality (Figure 3a), text unimodality (Figure 3b), and image-text multimodality (Figure 3c). We further developed DamageArbiter, a disagreement-driven arbitration framework that leverages the complementary strengths of both unimodal and multimodal models for this task (Figure 3d). After completing the model evaluation, we applied the best-performing architecture to street-view imagery collected after Hurricane Milton to assess the geographic distribution of disaster damage across various damage levels.

\begin{figure}[htbp]
\centering
\includegraphics[width=0.90\textwidth]{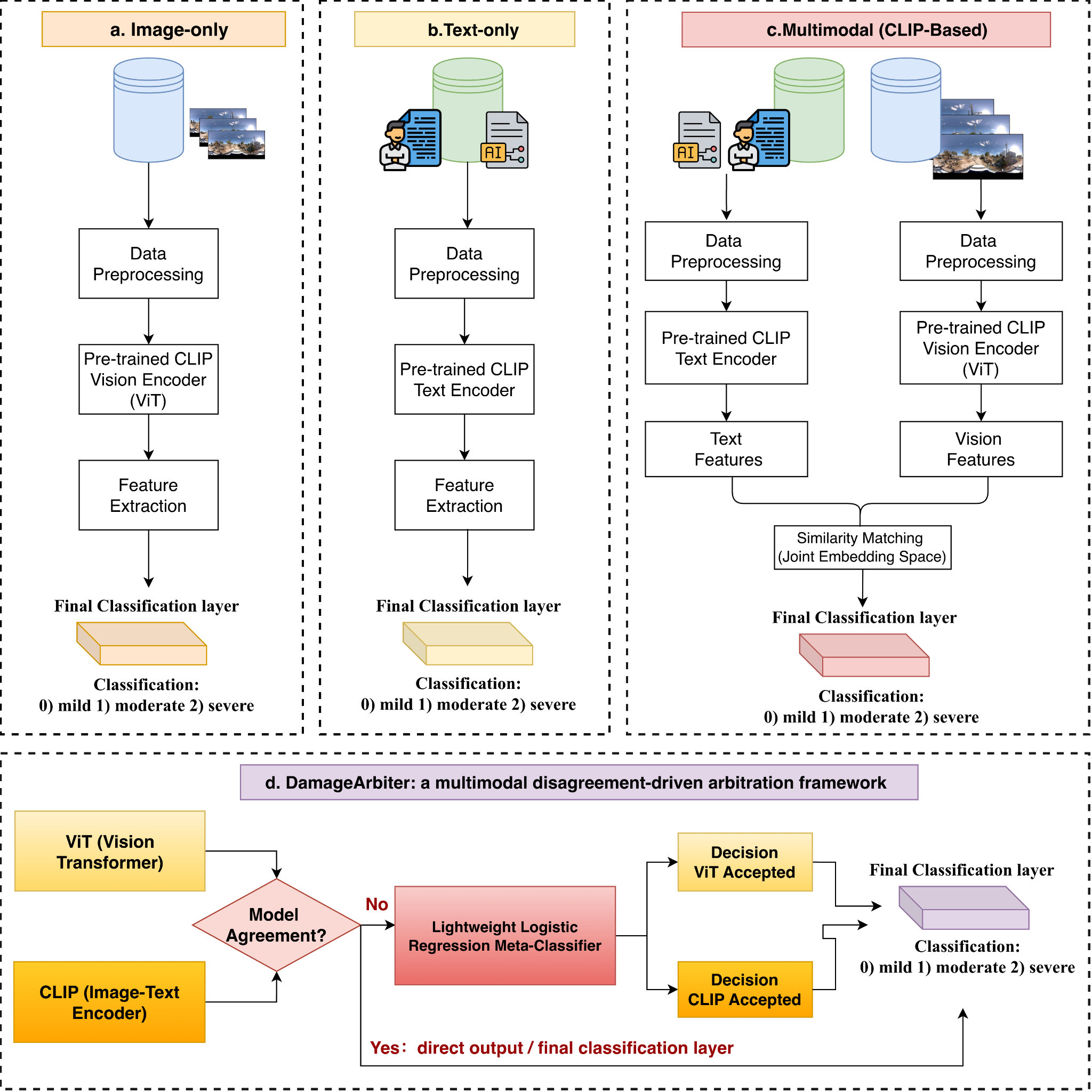}
\caption{Overall framework of disaster assessment approaches: (a) image-only model, (b) text-only model, (c) multimodal CLIP model, and (d) DamageArbiter, a multimodal disagreement-driven arbitration framework.}
\label{fig:framework}
\end{figure}

Section 4.2 presents the baseline model architectures. Section 4.3 describes DamageArbiter, including arbitration features, logistic regression training, and ablation design. Section 4.4 defines the evaluation metrics used to compare models' classification performance, semantic alignment, and reliability. All experiments were implemented in PyTorch using pre-trained models from the timm (\url{https://huggingface.co/timm}) and CLIP libraries. To ensure a fair comparison between unimodal and multimodal settings, all models were trained and evaluated using consistent data splits and hyperparameter configurations unless otherwise stated.

\subsection{Baseline Model Architectures and Implementation}

\subsubsection{Unimodal Image-Only Models}

We adopted the Vision Transformer (ViT) as the baseline unimodal image-only model for two reasons. First, as a Transformer-based architecture, ViT is designed to model global contextual relationships in images. This capacity is suitable for disaster scenes characterized by spatially distributed and heterogeneous visual patterns. Second, the visual encoder of CLIP is built upon ViT. Using ViT as the unimodal baseline ensures architectural consistency in the experimental design and model comparison.

The ViT model architecture is shown in Figure 4. All street-view images were uniformly scaled to $224\times224$ resolution and normalized before being input into the model. Data augmentation operations, such as random horizontal flipping, were applied during training to improve model generalization. Subsequently, the images were divided into fixed-size patches, which were transformed into sequential inputs via linear projection and position embedding, and then fed into a multi-layer Transformer Encoder. Each encoding layer consisted of multi-head self-attention, a feed-forward network, and layer normalization to capture global dependencies. Finally, the classification token (CLS) output was passed through a multi-layer perceptron (MLP) classification head to achieve a three-category prediction of disaster damage severity.

\begin{figure}[htbp]
\centering
\includegraphics[width=0.95\textwidth]{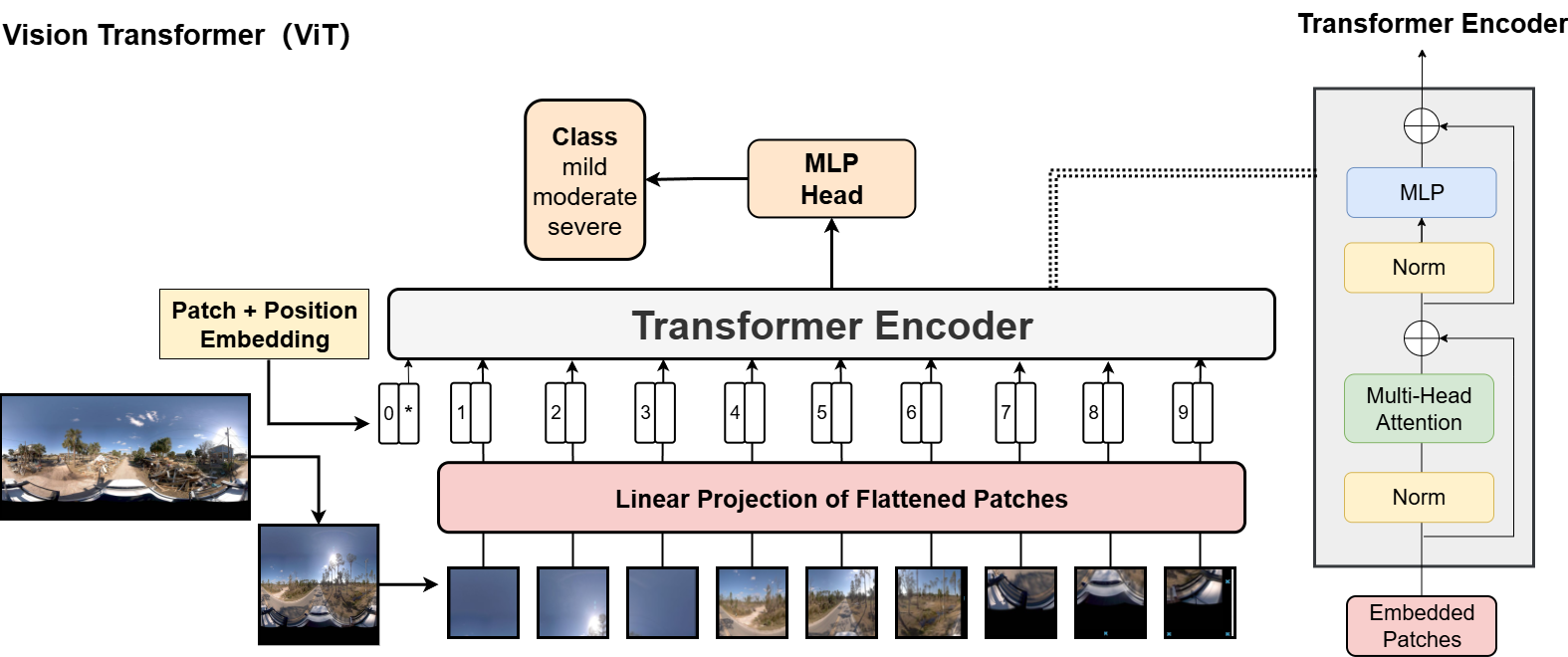}
\caption{Architecture of the unimodal image-only Vision Transformer (ViT) model for street-view-based disaster damage assessment}
\label{fig:vit-architecture}
\end{figure}

We implemented the ViT-Base-Patch16-224 and ViT-Base-Patch32-224 architectures with pre-trained weights from the timm library using PyTorch \citep{ref30}. The ViT baseline models were fine-tuned using the Milton-SV dataset. The AdamW optimizer was used for parameter updates, with a learning rate of $1\times10^{-4}$, weight decay of $1\times10^{-4}$, and a batch size of 32. Cross-entropy loss was used as the objective function, and the maximum number of training epochs was set to 10. To ensure model robustness, we employed a three-fold cross-validation, stratifying the entire dataset according to its class distribution. In each fold iteration, the training set comprised two-thirds of the dataset, and the validation set comprised the other one-third. Out-of-fold (OOF) predictions were also saved for subsequent overconfidence analyses.

\subsubsection{Unimodal Text-Only Models}

The text-only baseline experiment used CLIP's text encoder as a standalone classifier to isolate and evaluate the contribution of the text modality to disaster damage assessment. We set up two variants. The first, CLIP-Text Encoder (LLM captions), uses disaster descriptions automatically generated by GPT-4o-mini. These descriptions typically have broad, abstract semantic representations that summarize the overall characteristics of the disaster scene. The second, CLIP-Text Encoder (Human captions), uses descriptions written by trained human annotators. These descriptions focus on human-observable, fine-grained disaster signs, such as fallen trees, building debris, and damaged infrastructure. Experiments with LLM-generated input are conducted to test whether synthetic language can serve as an effective input for disaster damage assessment.

In implementation, we used the CLIP text encoder (Transformer architecture, 12-layer encoder, 512-dimensional word embeddings, maximum sequence length of 77) to convert each input text into a 512-dimensional semantic vector, which was then mapped to a three-category space (mild, moderate, and severe) via a linear classifier, as shown in Figure 5. The classification head was trained from scratch on the disaster dataset, while the CLIP text encoder was fine-tuned from pre-trained weights. Both text-only models used the same tokenization and encoding process and were evaluated with three-fold stratified cross-validation under identical hyperparameter settings (learning rate $1\times10^{-4}$, batch size 32, and AdamW optimizer).

\begin{figure}[htbp]
\centering
\includegraphics[width=0.95\textwidth]{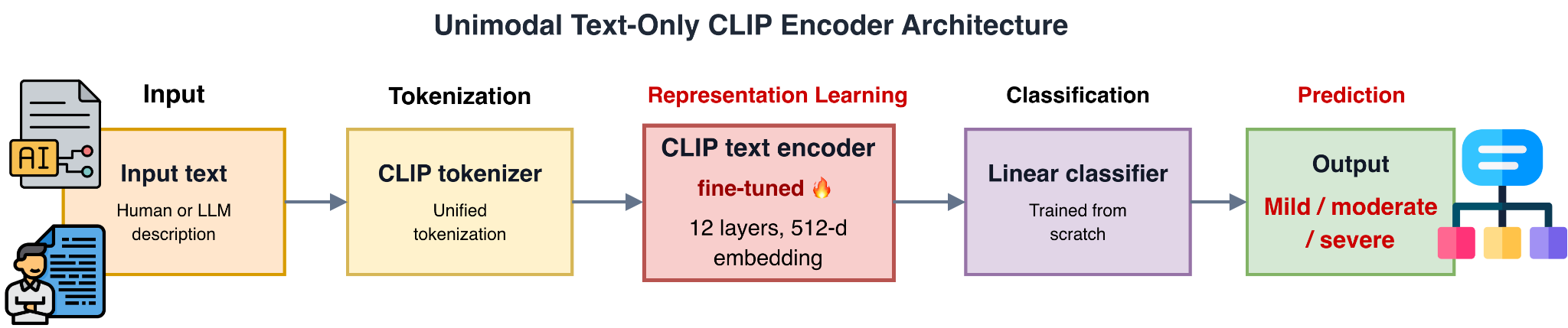}
\caption{Architecture of the unimodal text-only CLIP encoder for disaster severity classification.}
\label{fig:text-encoder}
\end{figure}

\subsubsection{Multimodal CLIP Models}

To learn robust multimodal representations for disaster assessment, we employed a cross-modal contrastive learning framework based on CLIP. Compared with visual-only or textual-only models, this cross-modal approach can improve interpretability and robustness in complex scenes with class imbalance and ambiguous visual cues. The CLIP-based model architecture is shown in Figure 6. Each fine-tuning sample consists of a post-disaster street-view image and its corresponding text description, either from human annotators or from GPT-4o-mini. This design allows us to compare how human-written and LLM-generated descriptions affect multimodal model learning and disaster assessment.

\begin{figure}[htbp]
\centering
\includegraphics[width=0.90\textwidth]{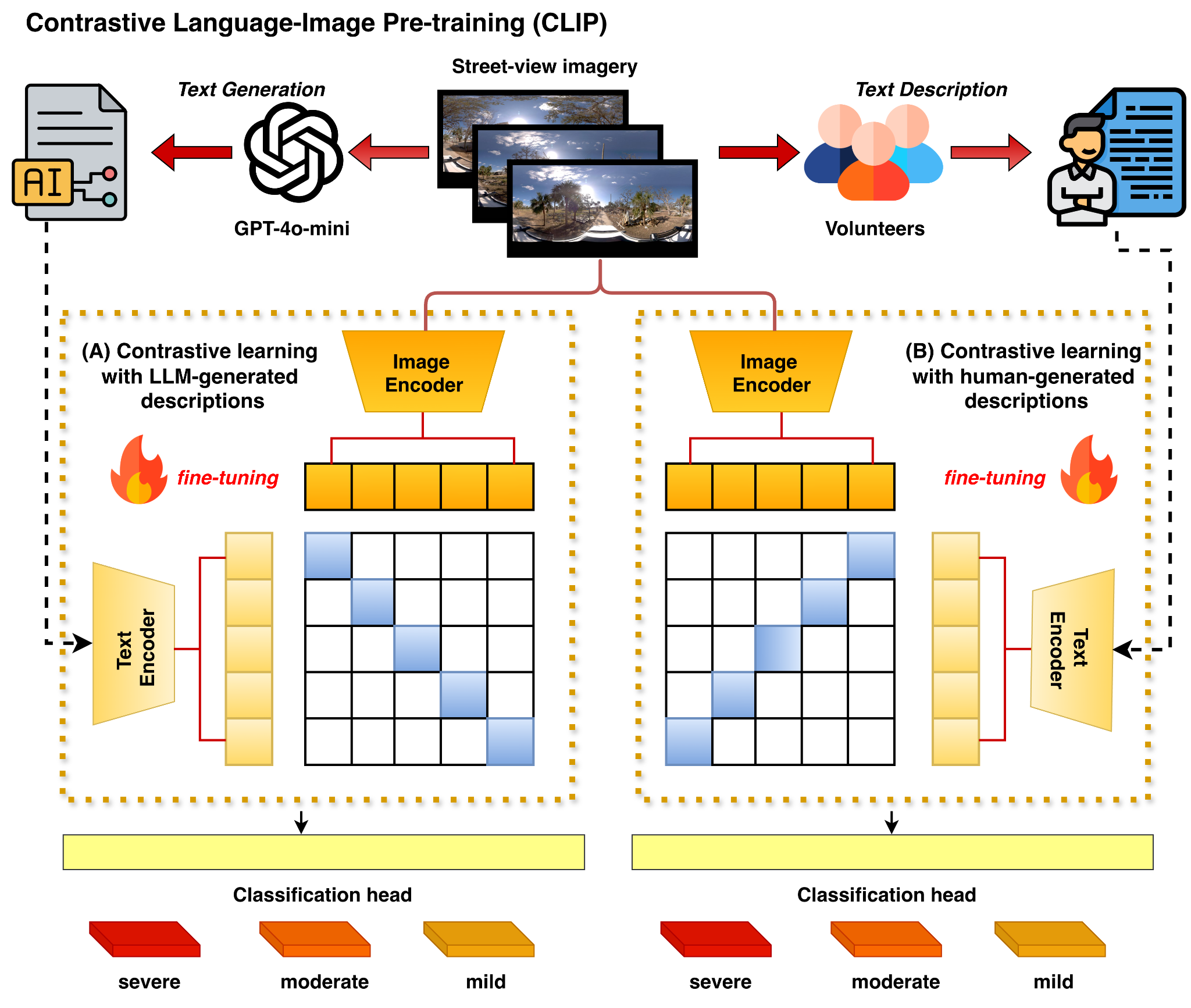}
\caption{CLIP-based contrastive learning for disaster assessment with human and LLM-generated descriptions.}
\label{fig:clip-contrastive}
\end{figure}

Visual and textual features are extracted using a pre-trained CLIP model, in which the image is processed by a visual encoder (ViT-B/16 as described in Section 4.2.1) and the text by a transformer-based text encoder (same as the encoder in Section 4.2.2). The resulting embeddings are projected into a joint representation space where cross-modal alignment can be learned through fine-tuning. During training, we optimized a single-directional InfoNCE loss, where each image serves as the query and its associated text description as the positive key. The objective encourages the model to increase similarity between matched image-text pairs while minimizing similarity between mismatched pairs within a batch. This is formulated as:

\begin{equation}
\mathcal{L}_{\mathrm{contrast}} = -\frac{1}{N}\sum_{i=1}^{N}\log
\frac{\exp\left(\mathrm{sim}(\mathbf{x}_i,\mathbf{y}_i)/\tau\right)}
{\sum_{j=1}^{N}\exp\left(\mathrm{sim}(\mathbf{x}_i,\mathbf{y}_j)/\tau\right)} .
\label{eq:contrastive-loss}
\end{equation}

where $\mathbf{x}_i$ and $\mathbf{y}_i$ denote the image and text embeddings of the $i$-th sample, $\mathrm{sim}(\mathbf{x},\mathbf{y})$ represents cosine similarity between two normalized embeddings, and $\tau$ is a temperature scaling factor. Here, $\tau$ is a learnable temperature inherited from the pre-trained CLIP model \citep{ref24} and fine-tuned jointly with the encoders, rather than a manually set hyperparameter.

To support the downstream classification, we appended a lightweight linear classification head to the combined embedding of the image and text modalities. This head is trained to predict one of three disaster severity levels (mild, moderate, or severe) using the standard cross-entropy loss:

\begin{equation}
\mathcal{L}_{\mathrm{cls}} = -\sum_{i=1}^{N}\sum_{c=1}^{C} y_{ic}\log \hat{y}_{ic} .
\label{eq:classification-loss}
\end{equation}

where $C$ is equal to 3 and denotes the number of severity classes, and $\hat{y}_{ic}$ is the predicted softmax probability for class $c$ of the $i$-th sample.

The final training objective combines the contrastive and classification losses as a weighted sum:

\begin{equation}
\mathcal{L}_{\mathrm{total}} = \lambda \mathcal{L}_{\mathrm{contrast}} + (1-\lambda)\mathcal{L}_{\mathrm{cls}} .
\label{eq:total-loss}
\end{equation}

with a tunable hyperparameter $\lambda \in [0,1]$ controlling the trade-off between representation alignment and classification accuracy. This hybrid training strategy enables the model to produce representations that are both semantically aligned across modalities and effective for downstream tasks such as damage severity prediction. This cross-modal contrastive learning framework is designed to improve the robustness of disaster assessment models. It also provides a principled setting for evaluating human-written and LLM-generated descriptions, especially when large batches of semantic descriptions are needed for rapid disaster response.

\subsection{DamageArbiter: A Disagreement-Driven Arbitration Framework}

\subsubsection{DamageArbiter Model Design}

DamageArbiter operates on the outputs of two fine-tuned baseline models, the ViT model described in Section 4.2.1 and the CLIP-LLM model described in Section 4.2.3. As shown in Figure 7, DamageArbiter uses a disagreement-driven arbitration strategy rather than directly averaging the ViT and CLIP-LLM predictions. For each test instance, the fine-tuned ViT and CLIP-LLM models first generate independent damage-category predictions. When the two models predict the same category, DamageArbiter uses this shared output as the final prediction. When the two base models disagree, a logistic regression arbitrator is activated to determine which model prediction should be adopted.

\begin{figure}[htbp]
\centering
\includegraphics[width=0.95\textwidth]{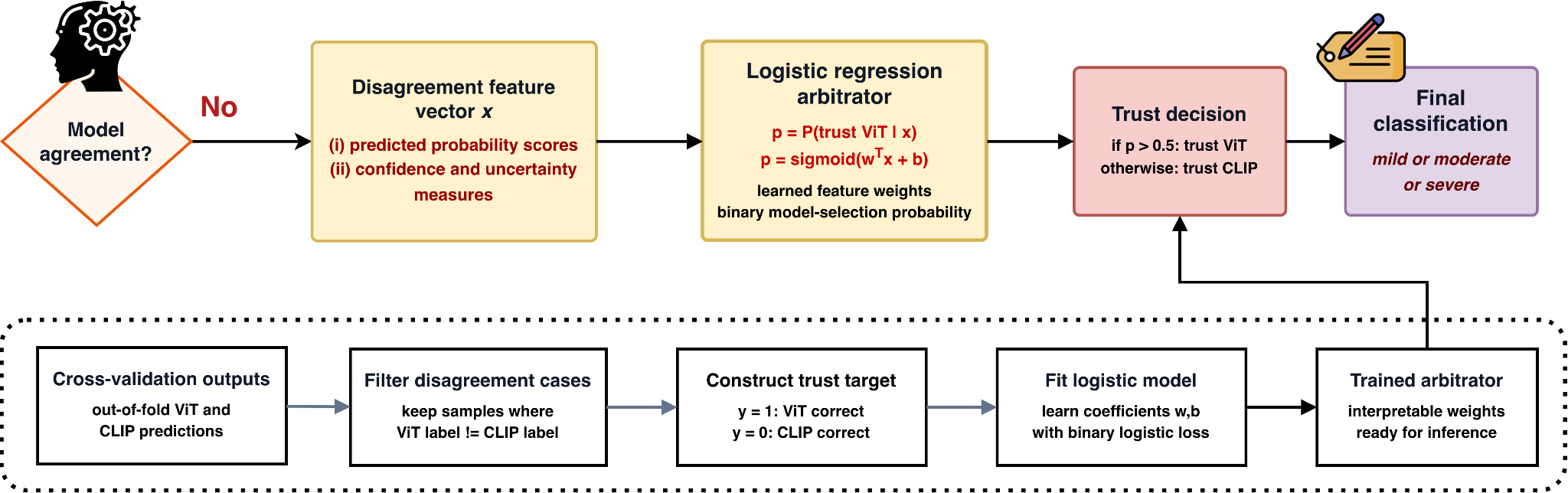}
\caption{Logistic regression-based DamageArbiter for disagreement-driven ViT-CLIP arbitration.}
\label{fig:damagearbiter}
\end{figure}

\subsubsection{Predictive Confidence and Uncertainty Measures}

We characterized each model's prediction using four measures derived from its predicted probability distribution over the three damage categories: two confidence-based error types (overconfidence and ambiguity) and two continuous uncertainty metrics (Shannon entropy and decision margin). Together, these measures describe not only how uncertain a prediction is, but also whether an incorrect prediction stems from misplaced confidence or from poor class separability.

The two confidence-based error types characterize the confidence level of incorrect predictions and were estimated through an analysis of misclassified samples (Figure 8). For each misclassified sample, we computed the confidence margin as the difference between the model's predicted probability for the incorrectly predicted class and its predicted probability for the ground truth class. A high confidence margin ($0.4$--$1.0$) indicates that the model was strongly confident in an incorrect prediction; such samples were labeled overconfidence errors. A low confidence margin ($0$--$0.1$) indicates that the predicted probabilities of the incorrect and true classes were nearly equal; such samples were labeled ambiguity errors. The remaining misclassified samples (confidence margin $0.1$--$0.4$) were categorized as medium-confidence errors. From a disaster-assessment perspective, overconfidence errors are particularly problematic because they reflect false certainty in incorrect predictions, whereas ambiguity errors are informative because they reveal cases of low-class separability, which may correspond to visually ambiguous damage patterns or insufficient evidence.

\begin{figure}[htbp]
\centering
\includegraphics[width=0.95\textwidth]{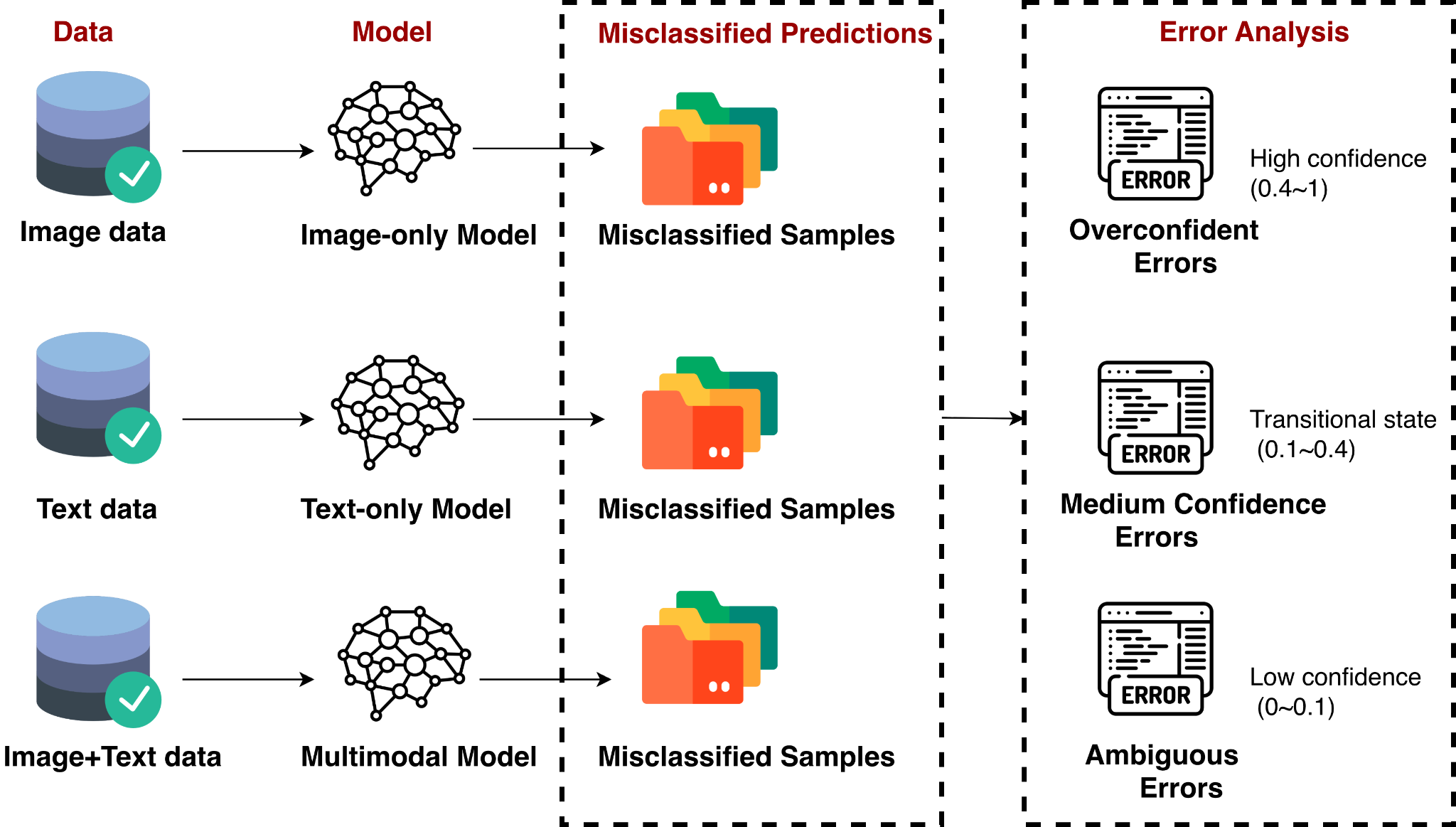}
\caption{Framework of confidence-based error analysis in disaster assessment models.}
\label{fig:confidence-error-analysis}
\end{figure}

The two continuous uncertainty metrics, entropy and decision margin, are computed directly from the full predicted probability distribution and therefore apply to every prediction, not only to misclassified samples.

Entropy quantifies how spread out the probability mass is across categories and is calculated as follows. Let $\mathbf{p}=(p_1,p_2,\ldots,p_C)$ denote a model's predicted probability distribution over the $C$ damage categories (here $C=3$), where $p_i$ is the predicted probability of category $i$ and $\sum_i p_i=1$. The Shannon entropy of this distribution is:

\begin{equation}
H(\mathbf{p}) = -\sum_{i=1}^{C} p_i \log p_i .
\label{eq:entropy}
\end{equation}

Entropy reaches its minimum ($H=0$) when the model assigns all probability to a single category (a fully confident, one-hot prediction), and its maximum ($H=\log C$) when the predicted probabilities are uniformly distributed across the $C$ categories (33\% probability for each category in this study). Higher entropy, therefore, indicates greater predictive uncertainty. We used the natural logarithm in this experiment, and any fixed logarithm base only rescales $H$ by a constant factor.

The decision margin is calculated as follows. Let $p_{(1)}$ and $p_{(2)}$ denote the highest and second-highest predicted probabilities for a given sample (i.e., the top-1 and top-2 class probabilities). The decision margin is their difference:

\begin{equation}
m = p_{(1)} - p_{(2)} .
\label{eq:decision-margin}
\end{equation}

A small margin means the two most likely categories are nearly tied, indicating weak class separability and high prediction ambiguity; a large margin means the top category is separated from the rest, indicating a confident, well-separated decision.

\subsubsection{Logistic Regression Training and Feature Selection}

Logistic regression was selected because it provides a lightweight and reproducible arbitration model suitable for the relatively small set of disagreement samples. The logistic regression arbitrator was trained using disagreement cases derived from the cross-validation outputs of the ViT and CLIP-LLM baseline models. For each fold, predictions were generated by models not trained on the corresponding validation samples, ensuring that the arbitrator was trained on out-of-fold predictions. We then identified validation samples for which ViT and CLIP-LLM produced different damage level predictions. The arbitration target was defined as a binary label. Samples were labeled as (y = 1) when the ViT prediction matched the ground truth and labeled as (y = 0) when the CLIP-LLM prediction matched the ground truth. Disagreement samples for which neither base model predicts the correct label were excluded from arbitrator training.

Given a disagreement feature vector $\mathbf{x}$, the logistic regression model estimates the probability that the final decision should trust ViT rather than CLIP as follows:

\begin{equation}
P(y=1\mid \mathbf{x}) = \sigma(\mathbf{w}^{\top}\mathbf{x}+b) .
\label{eq:logistic-arbitrator}
\end{equation}

where $\sigma$ denotes the sigmoid function, $\mathbf{w}$ is the coefficient vector, and $b$ is the intercept. At inference time,

let $g(\mathbf{x})=P(y=1\mid\mathbf{x})$ be the probability produced by the logistic regression arbitrator, where $\mathbf{x}$ is the arbitration feature vector and $y=1$ denotes that the ViT prediction should be trusted. When ViT and CLIP-LLM disagree, the final prediction follows ViT if $g(\mathbf{x})>\tau$ and follows CLIP otherwise, where $\tau$ is the arbitrator's decision threshold.

To identify the final feature set for the arbitrator, we conducted ablation experiments using progressively richer feature configurations. The first configuration (LOGREG\_conf) uses only each model's max-softmax confidence, that is, the probability assigned to its predicted class --- the single largest value of the three-class softmax distribution; the probabilities of the remaining two damage classes are not used as separate features. The second configuration (LOGREG\_conf+unc) incrementally adds two uncertainty measures, entropy and decision margin, both computed from the full three-class probability distribution and therefore summarizing information from all three classes rather than the predicted class alone. As a deployable reference, we also include a non-learned heuristic that selects the prediction of the more confident model.

All feature configurations were evaluated on the same selected disagreement samples and binary trust-label definition described above. For each configuration, a logistic regression arbitrator was trained using only features that can be computed without ground-truth labels at inference, namely predicted-class confidence, entropy, and decision margin. The decision threshold was not tuned: it was predefined and fixed at $\tau = 0.5$ for all configurations, so that the arbitrator adopts the ViT prediction when $g(\mathbf{x}) > 0.5$ and the CLIP prediction otherwise. Holding $\tau$ fixed avoids per-configuration threshold tuning, isolates the contribution of each feature group, and enables a fair comparison. After training, the logistic regression coefficients and the fixed threshold can be frozen and applied to newly collected, unlabeled street-view imagery using the same feature-computation procedure. Together, these ablation experiments quantify the contribution of confidence and uncertainty features to disagreement-driven arbitration and provide the empirical basis for selecting the optimal DamageArbiter configuration.

\subsection{Evaluation Metrics}

We evaluated the tested models in two steps. First, the baseline models were evaluated and compared using standard classification-based metrics and further analyzed using reliability measures to characterize confidence-based error patterns. In this step, image-text semantic alignment was also evaluated to assess the consistency between human- and LLM-generated annotations and the corresponding post-disaster street-view images. Second, the optimal DamageArbiter model configuration, identified through ablation experiments, was compared with the baseline models in terms of both classification accuracy and reliability.

Classification performance was evaluated using Accuracy, Recall, Precision, sample-weighted F1-score (SW-F1), and Matthews Correlation Coefficient (MCC), which were computed based on the confusion matrix components: true positives (TP), false positives (FP), true negatives (TN), and false negatives (FN). The classification metrics were computed using Equations~\eqref{eq:accuracy}--\eqref{eq:mcc}. SW-F1 was included to account for class imbalance across damage categories, whereas MCC was used as a balanced metric that considers all components of the confusion matrix.

\begin{align}
\mathrm{Accuracy} &= \frac{TP+TN}{TP+TN+FP+FN}, \label{eq:accuracy}\\
\mathrm{Recall} &= \frac{TP}{TP+FN}, \label{eq:recall}\\
\mathrm{Precision} &= \frac{TP}{TP+FP}, \label{eq:precision}\\
\mathrm{SW\mbox{-}F1} &= 2\times\frac{\mathrm{Precision}\times\mathrm{Recall}}
{\mathrm{Precision}+\mathrm{Recall}}, \label{eq:swf1}\\
\mathrm{MCC} &= \frac{TP\cdot TN-FP\cdot FN}
{\sqrt{(TP+FP)(TP+FN)(TN+FP)(TN+FN)}} . \label{eq:mcc}
\end{align}

Semantic alignment was evaluated using CLIPScore, a reference-free image-text alignment metric derived from CLIP. For each image-text pair, CLIP encodes the image ($I$) and text description ($T$) into embeddings $\mathbf{v}_I\in\mathbb{R}^d$ and $\mathbf{v}_T\in\mathbb{R}^d$, where $d$ is the dimensionality of the shared multimodal embedding space. CLIPScore was computed as the cosine similarity between the normalized image and text embeddings (Equation~\eqref{eq:clipscore}), with higher values indicating stronger semantic consistency. In this study, CLIPScore was used to compare the semantic alignment between human or LLM-generated annotations and the corresponding street-view images.

\begin{equation}
\mathrm{CLIPScore}(I,T) = \cos(\mathbf{v}_I,\mathbf{v}_T)
= \frac{\mathbf{v}_I\cdot\mathbf{v}_T}{\lVert\mathbf{v}_I\rVert\lVert\mathbf{v}_T\rVert} .
\label{eq:clipscore}
\end{equation}

Model uncertainty was evaluated using the confidence-based errors defined in Section 4.3.2, including overconfident, medium-confidence, and ambiguous errors.

DamageArbiter feature selection was evaluated through ablation experiments on disagreement samples. Each logistic regression arbitrator configuration was evaluated using Accuracy and MCC on the full test set. After the final feature configuration was selected, the resulting DamageArbiter model was evaluated against the baseline models using the same classification and confidence metrics, including Accuracy, Recall, Precision, SW-F1, MCC, over-confidence, medium-confidence, and ambiguous errors.

\section{Results}

\subsection{Baseline Model Performance and Reliability}

Table 1 summarizes the performance of all baseline models across five classification metrics, along with the CLIPScore values for the two baseline CLIP models. Overall, the image-only ViT-B/32 model achieved the strongest baseline performance across most metrics, including accuracy (0.7433), recall (0.7433), precision (0.7597), and MCC (0.5947). CLIP with human annotations achieved the highest SW-F1 score (0.7403), although its advantage over ViT-B/32 was marginal. The text-only baselines performed the worst across all five metrics, indicating that converting post-disaster images to text may omit visual cues that are important for distinguishing damage levels.

\begin{table}[htbp]
\centering
\caption{Comparative evaluation of image-only, text-only, and multimodal models on disaster severity classifications.}
\label{tab:baseline-performance}
\scriptsize
\resizebox{\textwidth}{!}{%
\begin{tabular}{llrrrrrrrrr}
\toprule
Model & Input modality & \multicolumn{5}{c}{Classification metrics} & \multicolumn{3}{c}{Reliability metrics} & CLIPScore \\
\cmidrule(lr){3-7}\cmidrule(lr){8-10}
 & & Accuracy & Recall & Precision & SW-F1 & MCC & Overconfident & Medium & Ambiguous & \\
\midrule
ViT-B/16 & Image only & 0.7261 & 0.7261 & 0.7320 & 0.7240 & 0.5653 & 64.86\% & 27.14\% & 8.00\% & -- \\
ViT-B/32 & Image only & \textbf{0.7433} & \textbf{0.7433} & \textbf{0.7597} & 0.7386 & \textbf{0.5947} & 70.58\% & 21.04\% & 8.38\% & -- \\
Text Encoder (LLM) & Text only & 0.6307 & 0.6307 & 0.6331 & 0.6261 & 0.4126 & 47.78\% & 37.71\% & 14.51\% & -- \\
Text Encoder (Human) & Text only & 0.6710 & 0.6710 & 0.6779 & 0.6715 & 0.4788 & 53.27\% & 33.65\% & 13.08\% & -- \\
CLIP (LLM) & Image+text & 0.7238 & 0.7238 & 0.7249 & 0.7229 & 0.5644 & 0.00\% & 55.18\% & 44.82\% & 0.2467 \\
CLIP (Human) & Image+text & 0.7422 & 0.7422 & 0.7504 & \textbf{0.7403} & 0.5915 & 0.00\% & 66.82\% & 33.18\% & 0.2701 \\
\bottomrule
\end{tabular}}
\par\smallskip
\parbox{\textwidth}{\footnotesize Note: Bold values indicate the best-performing model for each classification metric.}
\end{table}

For the multimodal CLIP baselines, CLIP-Human outperformed CLIP-LLM across all classification metrics, including accuracy (0.7422 vs. 0.7238), SW-F1 (0.7403 vs. 0.7229), and MCC (0.5915 vs. 0.5644). CLIP-Human also achieved a higher CLIPScore than CLIP-LLM (0.2701 vs. 0.2467), indicating stronger semantic alignment between human annotations and the corresponding street-view images. Consistently, the text-only CLIP model with human-written descriptions outperformed the version with LLM-generated descriptions. However, the performance gap between CLIP-Human and CLIP-LLM was relatively small, suggesting the potential of using LLM-generated annotations to enrich street-view imagery for disaster damage assessment when human annotations are limited, costly, or time-consuming to obtain. Nevertheless, CLIP-Human remained slightly below ViT-B/32 in accuracy and MCC, suggesting that the strongest baseline performance depended primarily on visual evidence.

The confidence-based error profiles in Table 1 reveal a different pattern. The image-only ViT models produced the highest proportions of overconfident errors, with 64.86\% of ViT-B/16 errors and 70.58\% of ViT-B/32 errors falling into the high-confidence error range. This indicates that the best-performing image-only baseline also produced the largest number of highly confident incorrect predictions. In contrast, the multimodal CLIP models produced no overconfident errors. Their errors were concentrated in the medium-confidence and ambiguous ranges, particularly for CLIP-LLM, for which ambiguous errors accounted for 44.82\% of all errors.

Overall, Table 1 shows that accuracy and reliability do not necessarily improve in parallel. Although ViT-B/32 achieved the best classification accuracy among the baseline models, it also produced the highest proportion of overconfident errors. In contrast, the multimodal CLIP models reduced overconfidence but still exhibited ambiguity and class-specific misclassification. These findings motivate the DamageArbiter framework, which uses disagreement-based arbitration to integrate the strong visual classification performance of ViT with the more conservative reliability profile of CLIP.

\subsection{Ablation Study and Performance Evaluation of DamageArbiter}

Table 2 summarizes the ablation experiment results. We ablated the arbitrator's feature set using label-free, inference-time features, at a fixed decision threshold $\tau = 0.5$. The accuracy on the full test set and the MCC scores were reported for each configuration. All learned configurations perform comparably (accuracy 0.7580-0.7585; MCC 0.6181-0.6188) and clearly exceed the non-learned heuristic that simply trusts the more confident model (0.7504; MCC 0.6063). The differences among learned configurations are within run-to-run variation: adding uncertainty features does not yield a meaningful improvement over the max-softmax confidence alone. We therefore selected the confidence-only configuration as the final arbitrator (accuracy 0.7585; MCC 0.6188).

\begin{table}[htbp]
\centering
\caption{Ablation study of logistic regression arbitration with different feature configurations.}
\label{tab:arbitration-ablation}
\begin{tabular}{lrr}
\toprule
Arbitration setting & Accuracy & MCC \\
\midrule
LOGREG\_conf & \textbf{0.7585} & \textbf{0.6188} \\
LOGREG\_conf+unc & 0.7580 & 0.6181 \\
Naive heuristic & 0.7504 & 0.6063 \\
\bottomrule
\end{tabular}
\end{table}

The ablation also clarifies how the arbitrator uses the two base models. Rather than applying a hand-set rule, the logistic-regression arbitrator is fit on the out-of-fold disagreement samples and learns, from the two models' max-softmax confidences, which prediction to trust on each divergent sample. Adding the uncertainty features (entropy and decision margin) does not improve accuracy or MCC over confidence alone (Table 2), so the final arbitrator uses confidence only. On the disagreement samples, this learned combination outperforms both the non-learned heuristic that always trusts the more confident model and either base model used in isolation. This behavior is consistent with the reliability profiles reported in Section 5.1, where the image-only model produced a high proportion of overconfident errors (70.58\%) while the multimodal CLIP model produced none (0\%): combining ViT's visual discrimination with CLIP's more conservative confidence yields more reliable decisions on the samples where the two models disagree.

Figure 9 summarizes the comparison of the two base models, the image-only ViT-B/32 and the multimodal CLIP-LLM, together with the proposed DamageArbiter, across the five classification metrics (panel a) and the three confidence-based error types (panel b). The classification metrics show that DamageArbiter outperforms both base models by improving the accuracy from 0.7433 to 0.7585, recall from 0.7433 to 0.7585, precision from 0.7597 to 0.7713, SW-F1 from 0.7386 to 0.7550, and MCC from 0.5947 to 0.6188. The largest improvement appears on the Matthews correlation coefficient (MCC), with an absolute gain of 0.0241 over ViT-B/32 and 0.0544 over CLIP-LLM. Because MCC accounts for all entries of the confusion matrix and is sensitive to class imbalance, this gain indicates that DamageArbiter produces more balanced and reliable predictions across damage categories, rather than merely improving performance for the majority class.

\begin{figure}[htbp]
\centering
\includegraphics[width=0.95\textwidth]{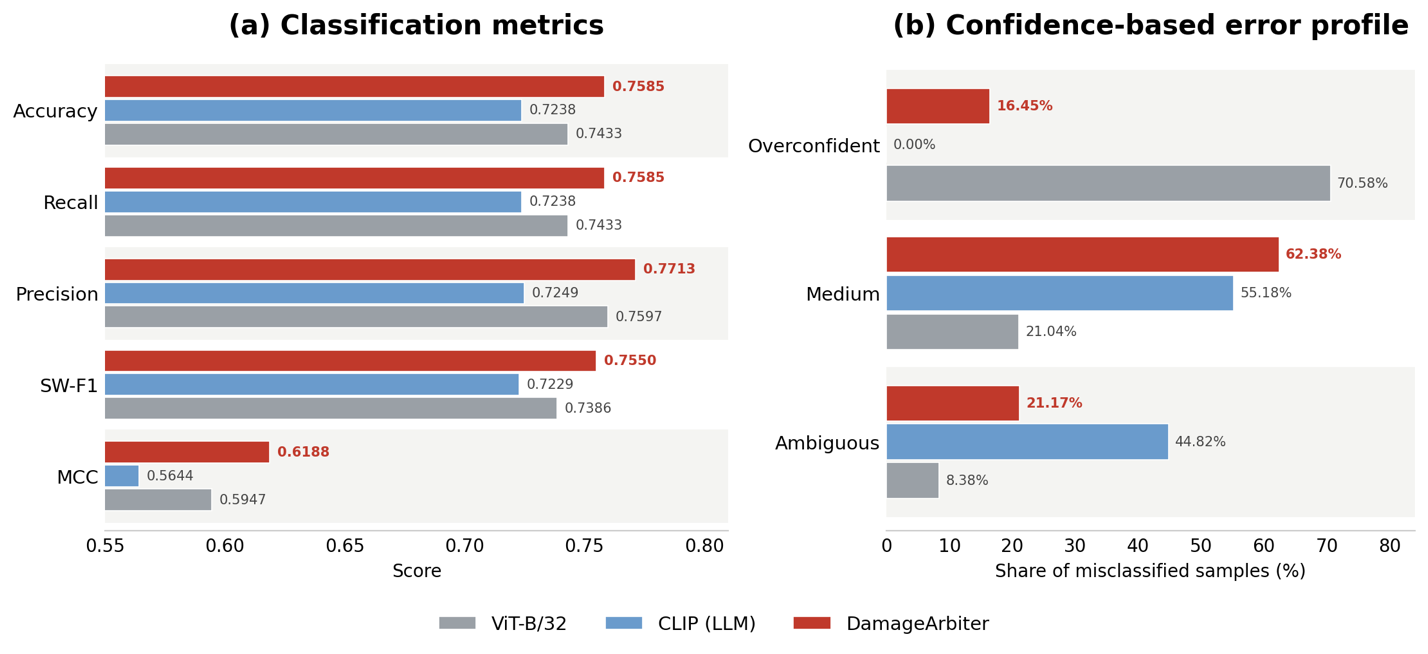}
\caption{Comparison of the two base models, ViT-B/32 (image-only) and CLIP-LLM (multimodal), and the proposed DamageArbiter. (a) Five classification metrics (higher is better); (b) the confidence-based error profile (overconfident, medium-confidence, and ambiguous error shares; lower overconfidence is better). DamageArbiter attains the highest score on every classification metric while sharply reducing the overconfident errors that dominate the image-only model.}
\label{fig:model-comparison}
\end{figure}

We further quantified the overconfidence profile of DamageArbiter by assigning each final prediction the probability distribution of the base model whose prediction was adopted. For the 2,041 agreement samples, where ViT-B/32 and CLIP-LLM predicted the same damage class, DamageArbiter retained the shared class prediction but used CLIP-LLM's more conservative probability distribution to characterize prediction confidence. For the disagreement samples, the confidence profile was determined by the base model selected by the logistic regression arbitrator. As shown in Figure~\ref{fig:model-comparison}(b), 16.45\% of DamageArbiter's errors were classified as overconfident, significantly lower than the ViT-B/32 baseline (70.58\%) and closer to the conservative profile of CLIP-LLM (0\%). This reduction reflects the structure of the arbitration framework. Because agreement samples account for most predictions, and because DamageArbiter assigns CLIP-LLM's more conservative probabilities to these cases, agreement errors no longer contribute to the high-confidence error category. The remaining overconfident errors arise primarily from disagreement cases in which the arbitrator selects the image-only ViT prediction. These results show that DamageArbiter improves classification accuracy while substantially mitigating the overconfident errors that dominate the image-only baseline. By combining ViT-B/32's strong visual discrimination with CLIP-LLM's more conservative confidence behavior, DamageArbiter provides a more reliability-aware prediction framework for rapid, hyperlocal disaster damage assessment from street-view imagery.

\subsection{Spatial Visualization}

We deployed the optimal DamageArbiter model to the Milton-SV data to examine the spatial structure of the predicted damage, the arbitration behavior, and the model's errors (Figure 10). The predicted-severity map (Figure 10b) closely reproduces the ground-truth pattern (Figure 10a), with the most severe damage concentrated along the exposed coastal and central streets in the study area. The arbitration-decision map (Figure 10c) shows where the disagreement-driven mechanism was engaged: the two base models agreed at the majority of locations (n = 2,041), whereas disagreements clustered along the main interior corridors, where the arbitrator retained the ViT prediction in 348 cases and adopted the CLIP prediction in 167. Finally, the error map (Figure 10d) shows that residual misclassifications are spread across the network rather than confined to one area. Consistent with the reduced overconfidence reported in Section 5.2, the residual overconfident errors highlighted in Figure 10d are sparse rather than dominant.

\begin{figure}[htbp]
\centering
\includegraphics[width=0.95\textwidth]{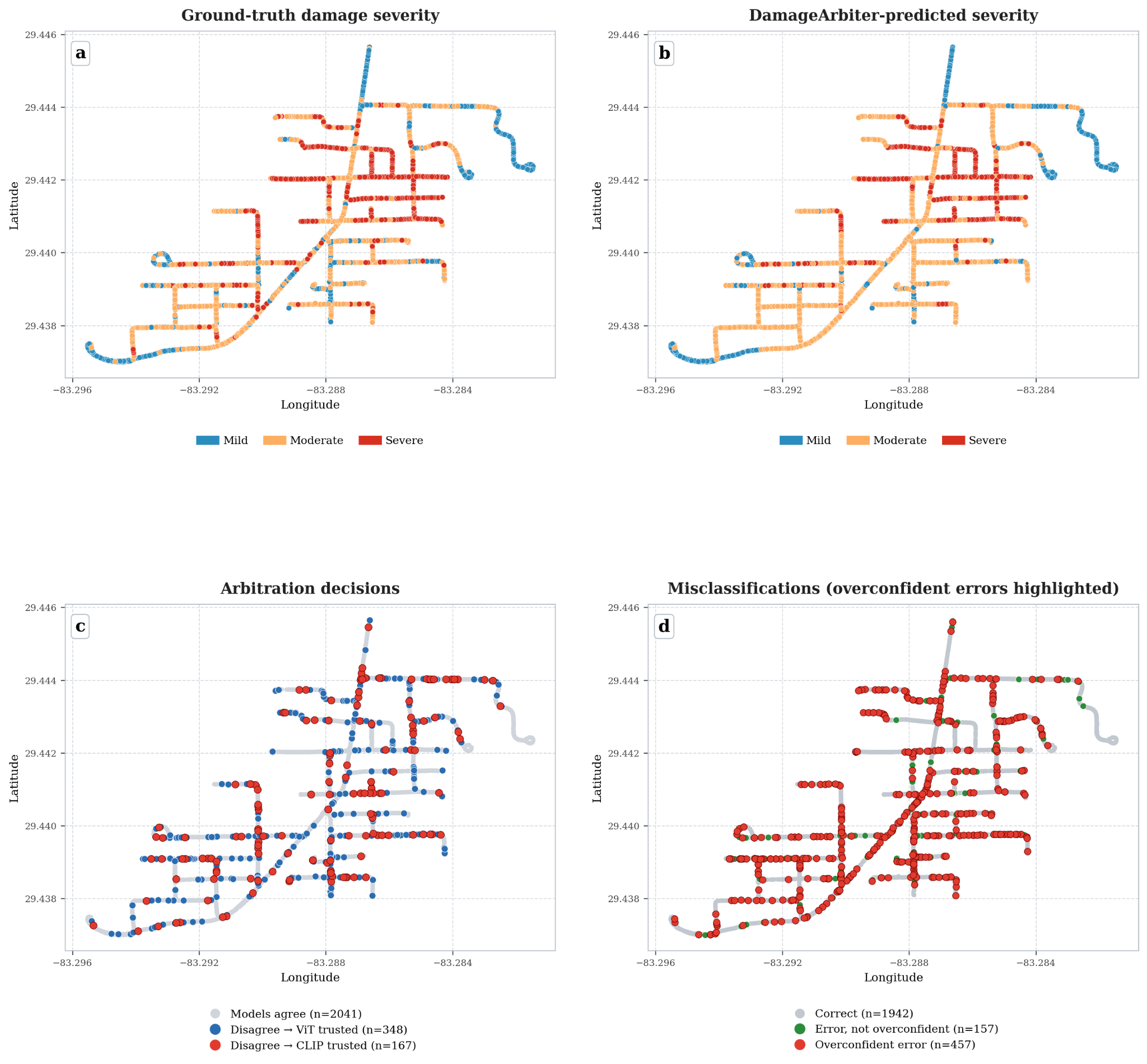}
\caption{Spatial deployment of DamageArbiter across street-view locations in Horseshoe Beach, Florida. (a) Ground-truth damage severity; (b) DamageArbiter-predicted damage severity; (c) arbitration outcomes; and (d) overconfident misclassifications by DamageArbiter.}
\label{fig:spatial-deployment}
\end{figure}

\section{Discussions}

This study proposes and evaluates DamageArbiter, a multimodal post-disaster assessment framework based on disagreement-driven arbitration. The results show that image-based models such as ViT provide strong visual discrimination for damage-severity classification, but their errors can be overconfident, whereas CLIP-based multimodal models show a more conservative reliability profile and provide a semantic interface for interpreting disaster-related visual cues. By using an arbitration strategy when ViT and CLIP disagree, DamageArbiter combines the discriminative strength of image-based classification with the semantic interpretability of language-image modeling, improving overall accuracy and balanced classification (MCC). Compared with the best baseline for each metric, DamageArbiter improved accuracy from 0.7433 to 0.7585, recall from 0.7433 to 0.7585, precision from 0.7597 to 0.7713, SW-F1 from 0.7403 to 0.7550, and MCC from 0.5947 to 0.6188, suggesting that disagreement-driven arbitration improves predictive performance and balanced classification. Crucially, because the arbitrator follows CLIP's more conservative confidence whenever ViT and CLIP agree, the overconfident-error rate of the DamageArbiter model falls to 16.45\%, far below the 70.58\% of the best-performing image-only baseline. This indicates that the arbitration strategy also substantially reduces the overconfidence inherited from the image model, thus improving model reliability.

The contributions of this work are threefold. First, it shows that accuracy alone is insufficient for evaluating classification-based damage-assessment models. An image-only model can attain competitive accuracy while committing a high proportion of overconfident errors (over 70\%). Hence, overconfidence should be quantified and reported alongside accuracy in image-based classification tasks like this case study. Second, it proposes DamageArbiter, a disagreement-driven framework that combines unimodal and multimodal models, improving accuracy and balanced classification (MCC) over the strongest baselines while making each model's overconfidence behavior explicit and interpretable through confidence analysis. Third, it shows that this overconfidence can be substantially reduced by the arbitration mechanism itself: by adopting CLIP's more conservative confidence when the two models agree, DamageArbiter lowers the proportion of overconfident errors from 70.58\% to 16.45\% without sacrificing accuracy.

Despite the findings, this study has two main limitations. First, the experiments focus on a single hurricane event and a limited geographic area, so the generalizability of the results to other regions, disaster types, and urban forms remains to be tested. This limitation is mainly due to the limited availability of high-quality post-disaster street-view imagery with reliable damage labels. The effectiveness of DamageArbiter across multiple hazards, geographic contexts, and built-environment settings needs to be further evaluated to better assess its transferability and robustness. Second, DamageArbiter is designed to assess damage by severity level (mild, moderate, and severe) rather than producing type-specific damage maps. Severity-level prediction is valuable for rapid disaster response because it can help identify the most heavily damaged areas and support prioritization of field inspection and resource allocation. In addition, the LLM-generated image descriptions used in this study can provide contextual information about local damage characteristics, thereby supporting more evidence-based interpretation of model outputs. Nevertheless, type-specific damage maps would provide more direct and operationally actionable information for emergency managers and infrastructure agencies.

These contributions and limitations point to three directions for future work. First, to move beyond the single-event, single-region scope, the framework should be evaluated and adapted across additional hurricane events, geographic regions, disaster types, and urban forms; where labeled post-disaster imagery is scarce, data augmentation, including synthetic post-disaster street-view generation, together with domain-adaptation techniques, could improve transferability. Second, to advance beyond severity-only assessment, the framework can be extended from severity classification toward damage-type mapping, distinguishing fallen trees, building debris, infrastructure disruption, and flooding, using open-vocabulary or zero-shot detection and segmentation with vision-language models, so that it produces the type-specific damage maps that operational response requires. Third, building on our finding that overconfidence can be reduced by the arbitration mechanism, future work should further integrate reliability into both model training and deployment. At the training stage, calibration strategies such as label smoothing or focal loss could be used to improve the reliability of predicted probabilities. At the inference stage, uncertainty-aware arbitration could explicitly incorporate confidence, entropy, and decision margin when determining which model prediction to trust. These confidence signals could also support human-in-the-loop deployment by automatically flagging low-confidence or high-disagreement assessments for field verification while allowing high-confidence predictions to support rapid situational awareness. These future research directions would extend DamageArbiter from a reliable severity-classification framework toward a transferable, type-aware, and operationally trustworthy disaster-assessment workflow.

\section{Conclusion}

This study develops DamageArbiter, a disagreement-driven multimodal arbitration framework for street-view-based post-disaster damage assessment. Using 2,556 post-disaster street-view images collected after Hurricane Milton in Horseshoe Beach, Florida, we systematically compared image-only ViT models, text-only CLIP models, multimodal CLIP models, and the proposed DamageArbiter framework for three-level damage severity classification. The results show that the image-only ViT-B/32 baseline model achieved the strong baseline accuracy but produced a higher proportion of overconfidence errors, whereas CLIP-based multimodal models exhibited a more conservative confidence profile but slightly lower classification performance. By arbitrating disagreements between the ViT-based model and CLIP-LLM, DamageArbiter achieved the highest overall performance, increasing accuracy to 75.85\% and MCC to 0.6188. More importantly, it reduced overconfident errors from 70.58\% for the image-only ViT-B/32 model to 16.45\%, demonstrating that disagreement-driven arbitration can improve prediction accuracy while substantially mitigating false certainty in model outputs.

Overall, this study makes two main contributions. First, it demonstrates that accuracy alone is insufficient for evaluating street-view-based disaster damage classification models and highlights the importance of incorporating confidence-based reliability measures, particularly overconfident errors, into model assessment. Second, it proposes and validates DamageArbiter as a lightweight and deployable arbitration framework that combines the strong visual discrimination of image-only models with the more conservative reliability profile of multimodal CLIP models. These findings support a more reliability-aware approach to AI-assisted disaster assessment from street-view data, especially in high-stakes contexts where confident but incorrect predictions may mislead response and recovery decisions. Future work should evaluate the framework across additional disaster types, geographic contexts, and urban forms; extend the current severity-level classification toward type-specific damage mapping; and further integrate calibration, uncertainty-aware arbitration, and human-in-the-loop verification into operational deployment.

\section*{Acknowledgements}

This article is based on work supported by three grants. One is from the U.S. National Science Foundation - Collaborative Research: HNDS-I: Cyberinfrastructure for Human Dynamics and Resilience Research (Award No. 2318206). The other two grants are from the Gulf Research Program under the U.S. National Academies of Sciences, Engineering, and Medicine (SCON-10000653; SCON-10001536). Any opinions, findings, conclusions, or recommendations expressed in this material are those of the authors and do not necessarily reflect the views of the funding agencies.

\section*{Declaration of Competing Interest}

No potential conflict of interest was reported by the authors.

\end{document}